\documentclass[sigconf,nonacm=true]{acmart} 
 
\AtBeginDocument{%
  }

 \setcopyright{none}
\copyrightyear{2023}
\acmYear{2023}

 \usepackage{balance}
\usepackage{subfigure}
\usepackage{multirow} 
\usepackage{booktabs} 
\usepackage{color}
\usepackage{bm}
\usepackage{mathrsfs}
\usepackage{makecell}
\usepackage{colortbl}
\usepackage{comment}
\usepackage{svg} 
\usepackage{caption}
\usepackage{enumitem}
\newcommand{\etal}{\textit{et al}.}
\begin{document}
 
\title{Feature Decoupling-Recycling Network for Fast Interactive Segmentation}

\author{Huimin Zeng}
\authornote{Both authors contributed equally to this research.}
\affiliation{
\institution{University of Science and Technology of China}
\city{Hefei}
\country{China}
}

\author{Weinong Wang}
\authornotemark[1]
\affiliation{%
  \institution{Xiaohongshu Inc}
 \city{Guangzhou}
\country{China}
}
 
\author{Xin Tao}
\affiliation{
\institution{Y-tech, Kuaishou Technology }
 \city{Shenzhen}
\country{China}
}
 
\author{Zhiwei Xiong}
\affiliation{%
  \institution{University of Science and Technology of China}
\city{Hefei}
\country{China}
}

\author{Yu-Wing Tai}
\affiliation{\institution{Dartmouth College}
\city{Hanover}
\country{United States}
}
 
\author{Wenjie Pei}
\authornote{Corresponding author. wenjiecoder@outlook.com}
\affiliation{
\institution{Harbin Institute of Technology}
\city{Shenzhen}
\country{China}
}

\renewcommand{\shortauthors}{Zeng et al.}

 \begin{abstract}
Recent interactive segmentation methods iteratively take source image, user guidance and previously predicted mask as the input without considering the invariant nature of the source image. As a result, extracting features from the source image is repeated in each interaction, resulting in substantial computational redundancy. In this work, we propose the \textbf{F}eature \textbf{D}ecoupling-\textbf{R}ecycling \textbf{N}etwork (\textbf{FDRN}), which decouples the modeling components based on their intrinsic discrepancies and then recycles components for each user interaction. Thus, the efficiency of the whole interactive process can be significantly improved. To be specific, we apply the Decoupling-Recycling strategy from three perspectives to address three types of discrepancies, respectively. First, our model decouples the learning of source image semantics from the encoding of user guidance to process two types of input domains separately. Second, \textbf{FDRN} decouples high-level and low-level features from stratified semantic representations to enhance feature learning. Third, during the encoding of user guidance, current user guidance is decoupled from historical guidance to highlight the effect of current user guidance.  We conduct extensive experiments on 6 datasets from different domains and modalities, which demonstrate the following merits of our model: 
1) superior efficiency than other methods, particularly advantageous in challenging scenarios requiring long-term interactions (up to 4.25x faster), while achieving favorable segmentation performance; 2) strong applicability to various methods serving as a universal enhancement technique; 3) well cross-task generalizability, \emph{e.g.,} to medical image segmentation, and robustness against misleading user guidance.
\end{abstract}

 
\maketitle

\section{Introduction}

Interactive segmentation has attracted great interest in computer vision area since decades ago, which introduces various kinds of user annotations (\emph{e.g.,}~point clicks~\cite{lin2020interactive,sofiiuk2021reviving,jang2019interactive, sofiiuk2020f}, strokes~\cite{bai2014error, boykov2001interactive, grady2006random, veksler2008star} and bounding boxes~\cite{rother2004grabcut, wu2014milcut, cheng2015densecut}) in segmentation process. Thus users can better express their intents and manually improve segmentation quality. It has served as a basic tool in image/video editing software, and achieve considerable improvements thanks to recent progress in deep learning.

 \begin{figure}[t]
    \centering
\includegraphics[width=0.8\linewidth]{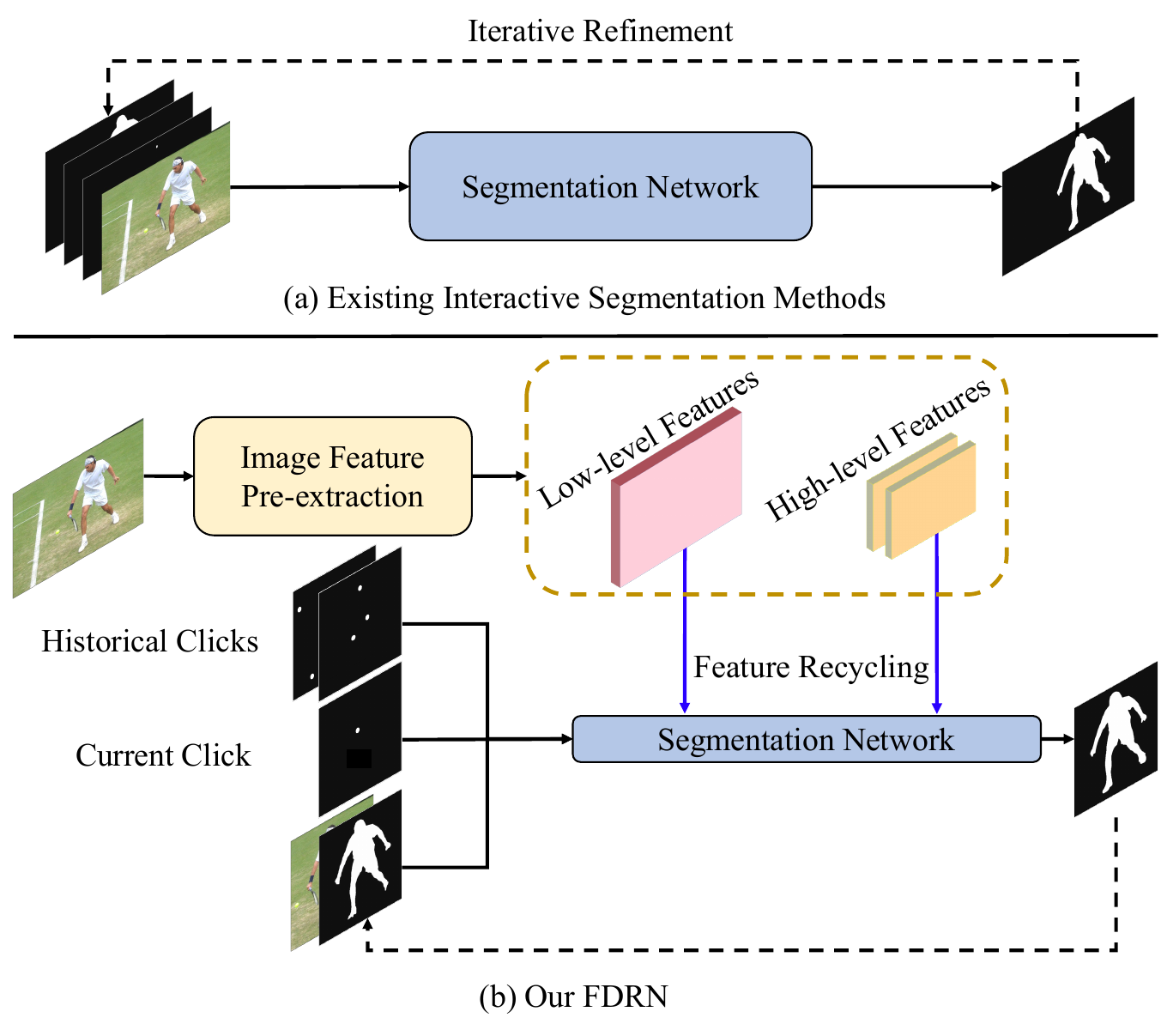}\vspace{-12pt}
    \caption{(a) Existing methods repeat the whole modeling process in each iteration; (b) Our \emph{FDRN} performs three decouplings, including the decoupling 1) between image features and user guidance; 2) between high-level and low-level features and 3) between current and historical user guidance, improving the modeling efficiency substantially.} 
     \label{fig:overview}
\end{figure}

While a stream of methods requires online-training~\cite{sofiiuk2020f,jang2019interactive}, most recent works can also achieve good results with feed-forward networks~\cite{sofiiuk2021reviving,focalclick}.
Representative approaches~\cite{sofiiuk2021reviving,focalclick,liupseudoclick} combine source images and user guidance with convolution layers, and then feed them into backbone networks recurrently for segmentation (see Fig.~\ref{fig:overview}(a)). Compared to conventional methods, the computation cost of these methods is more time sensitive in order to minimize the waiting time of users, which hinders the use of large backbones in practical applications. After investigating components of the interactive segmentation task, we believe that by delicately designed decoupling and recycling schemes, a more effective and efficient framework can be achieved. We first introduce our observations:

\noindent\textbf{Domain discrepancy. } The inputs of this task are from different domains, namely source images and interaction maps encoded from user guidance. While source images contain rich information and require a large backbone for effective feature extraction, user interaction maps are less informative with only spatial clues. Moreover, images remain constant and only user interaction maps change. Therefore, it is possible to decouple and separately process these two types of inputs, followed by a lightweight interaction stage, which may simultaneously achieve high quality and efficiency.
 
\noindent\textbf{Stratified semantic representations. } Both high-level and low-level features are of vital importance for accurate segmentation. While the former help understand the object of interest and propagate sparse interactions to long-range, the latter provide accurate boundaries and details. Due to their different nature, we propose to separately enhance these features with two specially designed modules, which has not been explored by previous methods.

\noindent\textbf{Temporal variance. } Most of the existing methods model this problem as a recurrent process. However, in each time step, a majority of them encode current user clicks and previous ones together, treating them equally. We notice that compared to historical clicks, the latest user click is more likely to indicate a previously misestimated area and gives clearer clues for improving accuracy. Therefore, we propose to decouple the encoding process of current clicks and historical clicks, emphasizing on the current click.

In this paper, we propose a novel framework named \emph{Feature Decoupling-Recycling Network} (\emph{FDRN}) for efficient interactive segmentation (as shown in Fig.~\ref{fig:overview}(b)). Corresponding to the observations above, our method performs three types of important decoupling with appropriate operations, including 1) decoupling between image features and user guidance; 2) decoupling between high-level and low-level semantic features, and 3) decoupling between current and historical user guidance. 
 
In summary, our contributions are three-fold:
\begin{itemize}[leftmargin =*, itemsep =0pt, topsep =0pt]
\item  We proposed a simple yet effective feature-decoupling-recycling framework named \emph{FDRN} for fast interactive image segmentation, which performs three types of decoupling and further recycles pre-extracted features during iterative interactions to reduce computational redundancy. 
\item Our method is a flexible and sufficiently general improvement scheme that can be easily integrated into a wide variety of networks and backbones.  
\item Experimental results show that compared to previous approaches, our method achieves significant speedups of up to $4.25\times$ without sacrificing performance, while showing strong generalization ability for medical images and robustness in practical scenarios.
\end{itemize}

\section{Related Work} 
\noindent\textbf{Optimization-based interactive image segmentation. }
Early works~\cite{boykov2001fast,rother2004grabcut,agarwala2004interactive} solve the interactive segmentation problem from the aspect of optimization.
Boykov and Jolly~\cite{boykov2001interactive} consider users-labeled pixels as hard constraints for predicted masks.
Afterwards~\cite{blake2004interactive,grady2006random} predict pixel-wise probability distribution and assign each pixel to the label of the maximum probability for final results.  
Gulshan \etal~\cite{gulshan2010geodesic} introduce multiple stars and geodesic path into shape constraint. Kim \etal~\cite{kim2010nonparametric} address quadratic cost functions of pixel and region and utilize long-range connections to facilitate regional cues propagation across larger domains.  

\noindent\textbf{Deep learning-based interactive image segmentation. }
Xu \etal~\cite{xu2016deep} explore to handle interactive segmentation with fully convolutional networks. RIS-Net~\cite{liew2017regional} expands the field-of-view of the given inputs to capture local regional information surrounding them for better local refinement. Li \etal~\cite{li2018interactive} adopts a selection network to choose the most plausible solution from a diverse set of outputs. 
Seednet~\cite{song2018seednet} proposes to learn an agent for interactive segmentation with deep reinforcement learning to better simulate the users' inputs.
MultiSeg~\cite{liew2019multiseg} incorporates scale priors to encourage diverse scales of results. FCA-Net~\cite{lin2020interactive} analyses the influence of the first click and proposes a click-guided loss function. Hu~\etal~\cite{hu2019fully} take a two-stream network to reduce interactions between user inputs and image features, thus helping current user input to have a higher impact on the results.   
 
BRS~\cite{jang2019interactive} combines optimization-based and deep learning-based approaches and develops a back-propagating refinement scheme (BRS) to modify the mislabeled regions. f-BRS~\cite{sofiiuk2020f} improves BRS~\cite{jang2019interactive} with partial backward propagation to accelerate the optimization process. These online-learning methods are slow during inference time due to the requirement of multiple optimizations. RITM~\cite{sofiiuk2021reviving} deals with click-based interactive segmentation as a simple feed-forward procedure. PseudoClick~\cite{liupseudoclick} proposes to simulate human inputs to reduce the user interaction cost and refine the segmentation mask.
FocalClick~\cite{focalclick} decomposes the mask prediction into coarse segmentation and local refinement, and proposes to update masks from a local view. 

Although methods mentioned above make great progress in interactive segmentation, they  need to refine results with the whole network, thus struggling to achieve the balance between performance and speed.  In contrast, our method performs image semantics extraction in the beginning and reuses them to avoid extracting features repeatedly, which significantly speeds up the interaction process.

\begin{figure*}[t]
\centering
\includegraphics[width=1\linewidth]{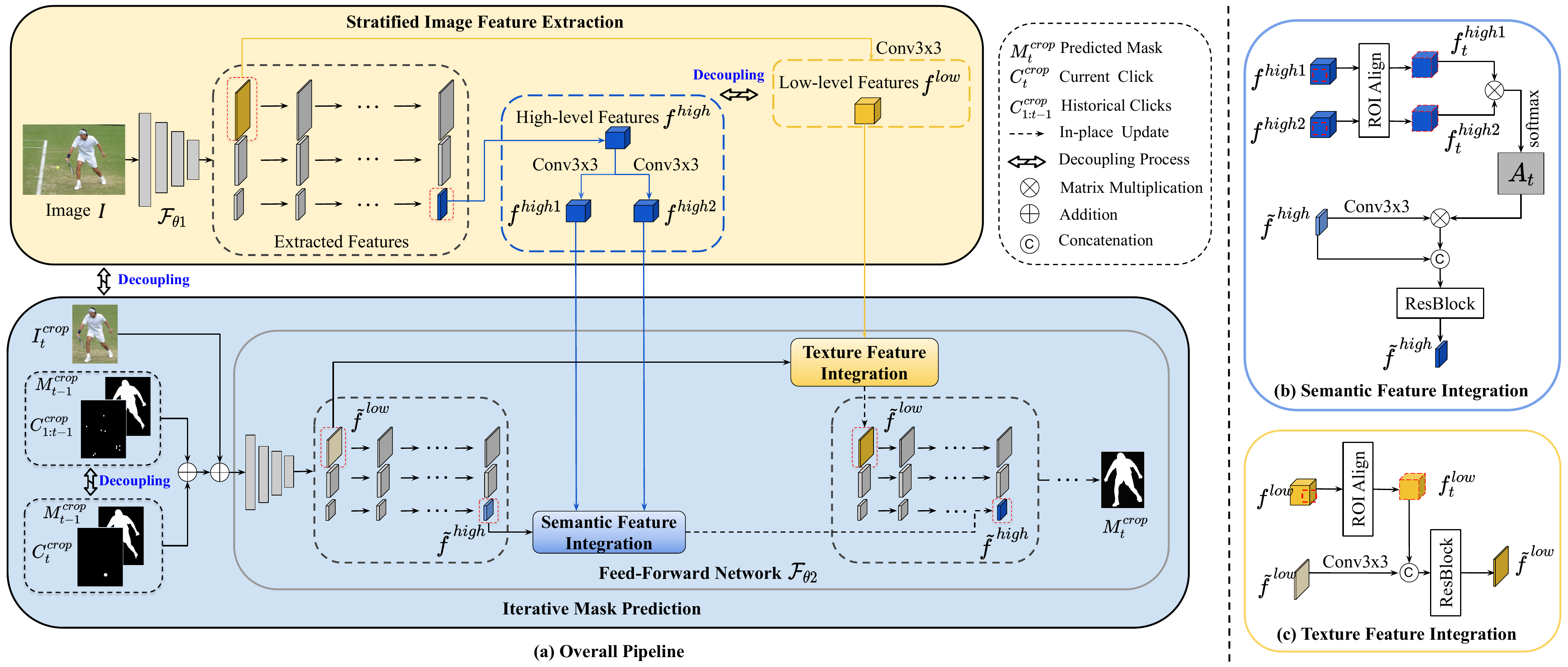}\vspace{-10pt}
\caption{Pipeline of the proposed \emph{FDRN}. \emph{FDRN} decouples the learning of image features and the encoding of user guidance by modeling them separately in two stages: 1) Stratified Image Feature Extraction, in which the high-level and low-level features are further decoupled, and 2) Iterative Mask Prediction, which recycles the image features from the first stage, and iteratively encodes the user guidance to predict segmentation mask. In this stage, the historical clicks and current click are decoupled to highlight the effect of current click.} 
\vspace{-8pt}
\label{global architecture}
\end{figure*}

\section{Method}
\subsection{Overview}
\noindent\textbf{Problem formulation. }\label{sec: Problem Formulation} We focus on the click-based interactive segmentation problem, which takes a source image $I$ and a series of user click locations $\{C_i\}$ as inputs, and outputs a binary segmentation mask $M$. Typical methods model the user interaction as an iterative process and output a series of masks $\{M_i\}$ sequentially. Formally, given a source image $I$, the segmentation model $\mathcal{F}_{\theta}$ predicts the mask in the $t$-th interaction based on previously predicted $M_{t-1}$ and user clicks $C_{1:t}$ in all steps with the expression as follows,
\vspace{-10pt}\begin{equation}\label{paradigm:v1} 
  M_t = \mathcal{F}_{\theta}(I,M_{t-1}, C_{1:t}).
\end{equation}
In contrast to existing methods~\cite{sofiiuk2021reviving,focalclick,liupseudoclick} that jointly learn semantic feature and encode user guidance, our \emph{FDRN} decouples them into two separate stages (see Fig.~\ref{global architecture}(a)).  Therefore, backbones of different complexity can be applied to these stages for high-quality image feature learning and user guidance encoding, respectively.

\noindent\textbf{Decoupling of image features and user guidance. } As illustrated in Fig.~\ref{global architecture}(a), the proposed \emph{FDRN} is a generic framework with two stages: 1) Stratified Image Feature Extraction and 2) Iterative Mask Prediction.   The first stage extracts stratified semantic features from source image $I$ only once, which can be regarded as initialization in practical applications. The second stage recycles the obtained semantic features and iteratively encodes user interactions to predict segmentation masks.  Therefore, such a decoupling-recycling strategy significantly elevates the efficiency of the interactive process. 

We first elaborate on these two stages in Sec.~\ref{sec: Image Semantics Extraction} and Sec.~\ref{sec: Mask prediction stage}, respectively, and then describe how to train \emph{FDRN} in an end-to-end manner with the proposed dynamic training strategy in Sec.~\ref{Dynamic-scale Strategy}.
 
\vspace{-4pt}\subsection{Stratified Image Feature Extraction}\label{sec: Image Semantics Extraction}
To extract high-quality image semantics from the source image,  we equip $\mathcal{F}_{\theta 1}$ with backbones that are suitable for the segmentation task and widely adopted by existing methods (\emph{e.g.,} HRNet~\cite{WangSCJDZLMTWLX19} and Segformer~\cite{xie2021segformer}). 
 
\noindent\textbf{Decoupling of high-level and low-level features. } Different from previous methods that stack multi-level features within a single network, we propose to decouple the stratified features into high-level and low-level features, and make use of them individually for different purposes. As shown in Fig.~\ref{global architecture}(a), we extract the features with the lowest resolution from relatively deep layers as high-level features (denoted as $f^{high}$) to capture the semantics of source image $I$. Then our \emph{FDRN} performs linear transformations on the high-level image features $f^{high}$ by individual convolution layers and obtains two views of the features, denoted as $f^{high1}$ and $f^{high2}$.  The obtained $\{f^{high1},f^{high2}\}$ are further fed into the Semantic Feature Integration sub-module for deep interaction with the encoded high-level features of user guidance, which will be explicated in Sec.~\ref{sec:recycling}. 
On the other hand, we extract low-level features (denoted as $f^{low}$) with the highest resolution from relatively shallow layers, which contain detailed texture information of the source image. We then integrate them with the low-level features of user guidance in the proposed Texture Feature Integration sub-module. We will explain it concretely in Sec.~\ref{sec:recycling}.

Decoupling the image features into high-level and low-level features allows our \emph{FDRN} to fully exploit the potentials of the learned image features and perform feature fusion with the encoded features of user guidance across different semantic levels, thereby achieving more precise segmentation results. Ablation studies in Sec.~\ref{sec:Ablation Studies} validate the merits of such design.

\subsection{Iterative Mask Prediction}\label{sec: Mask prediction stage}
In the stage of Iterative Mask Prediction, our \emph{FDRN} iteratively encodes the user guidance and incorporates the learned image features to predict the segmentation mask.  As shown in Fig.~\ref{global architecture}(a), \emph{FDRN} follows~\cite{sofiiuk2020f,focalclick,sofiiuk2021reviving} to crop a region of interest according to the outermost bounding box jointly determined by positive clicks and boundaries of the previously predicted mask.  Note that we leverage ROI Align~\cite{he2017mask} to obtain fixed-size feature maps for all these cropped inputs and image features.  Therefore, \emph{FDRN} can focus on the target region and get rid of distractions from interaction-irrelevant areas. The stage of Iterative Mask Prediction performs mask prediction with a feed-forward network $\mathcal{F}_{\theta 2}$, which is implemented with variants of HRNet~\cite{WangSCJDZLMTWLX19} and Segformer~\cite{xie2021segformer}  according to practical requirements for performance and inference time.

\noindent \textbf{Decoupling of current and historical user guidance. }
Compared with historical user guidance, current user guidance gives more related clues to the mislabeled area and is potentially more informative for refining the segmentation. Therefore, we propose to decouple current user guidance from historical guidance, and encode them with separate convolution layers during the encoding of user guidance. In this way, \emph{FDRN} can focus on utilizing current user guidance and highlight its effect on the prediction of segmentation. 
Specifically, our \emph{FDRN} first transforms the current and historical clicks into binary maps (denoted as $C_t^{crop}$ and $C_{1:t-1}^{crop}$), and then integrates each of them with the last predicted mask $M^{crop}_{t-1}$, respectively. Each of the integrated features is further fed into a separate Conv1S~\cite{sofiiuk2021reviving} layer for encoding. Finally, \emph{FDRN} fuses the encoded features of current and historical user guidance by performing convolutions on their concatenation.

\subsubsection{Recycling of Image Features}
\label{sec:recycling}
In the stage of feature extraction, the extracted image features are decoupled into high-level and low-level features.  In the stage of mask prediction, our \emph{FDRN} further recycles these features and integrates them with the encoded features of user guidance by the proposed Semantic Feature Interaction module and Texture Feature Interaction module, respectively.

\noindent\textbf{Semantic feature integration. } High-level image features typically contain semantic information of source image, which help propagate sparse user clicks to long-range areas. Therefore, we employ non-local attention~\cite{wang2018non} to enhance such connections, which is illustrated in Fig.~\ref{global architecture}(b). Specifically, our \emph{FDRN} first imposes ROI Align~\cite{he2017mask} on global semantic features $\{f^{high1},f^{high2}\}$ to obtain local fixed-size features $\{f_t^{high1},f_t^{high2}\}$. Then we reshape them into vectorial forms $\{K_t,Q_t\} \in \mathbb{R}^{hw\times C^k}$ and calculate pixel-pair-wise affinity matrix as $W_t\in \mathbb{R}^{h w \times hw}$, which is further softmax-normalized as $A_t$:
\vspace{-8pt}\begin{equation}
   W_t = K_tQ_t^{T}, \
\end{equation}
\vspace{-12pt}\begin{equation}
    A_t^{jk} = \frac{exp(W_t^{jk})}{\Sigma_{n}(exp(W_t^{nk}))}.
\end{equation}
Similar to the decoupling of high-level and low-level image features, we also decouple the encoded features of user guidance into high-level and low-level features, denoted as $\widetilde{f_t}^{high}$ and $\widetilde{f_t}^{low}$, respectively. Then we enhance the $\widetilde{f_t}^{high}$ using the obtained normalized affinity matrix $A_t$:
\vspace{-10pt}\begin{equation}
    X_t =  A_t Conv(\widetilde{f_t}^{high}),
     \label{reweight}
 \end{equation}
where $Conv$ denotes a $3\times 3$ convolutional operation. The obtained $X_t$ is then concatenated with original $\widetilde{f_t}^{high}$ and projected to original channel dimension via a ResBlock~\cite{he2016deep}. We substitute  $\widetilde{f_t}^{high}$ with such new feature, before being fed into the rest of the network.  Note that, although non-local attention is time consuming, we only apply it to high-level features with very low resolutions (\emph{e.g.,}~$\frac{h}{16}\times \frac{w}{16}$).
 
\noindent\textbf{Texture feature integration. } Low-level image features are rich in textures and details. As shown in Fig.~\ref{global architecture}(c), we simply concatenate low-level image features $f_t^{low}$ with low-level features of user guidance $\widetilde{f_t}^{low}$, then project back to the original dimension with a ResBlock~\cite{he2016deep} to replace the previous features. 
Such simple fusion operation allows Texture Feature Integration to perform efficiently. Experiments in Sec.~\ref{sec:Ablation Studies} validates the effectiveness of Texture Feature Integration.

\begin{figure}[t]
    \centering
    \includegraphics[width=0.9\linewidth]{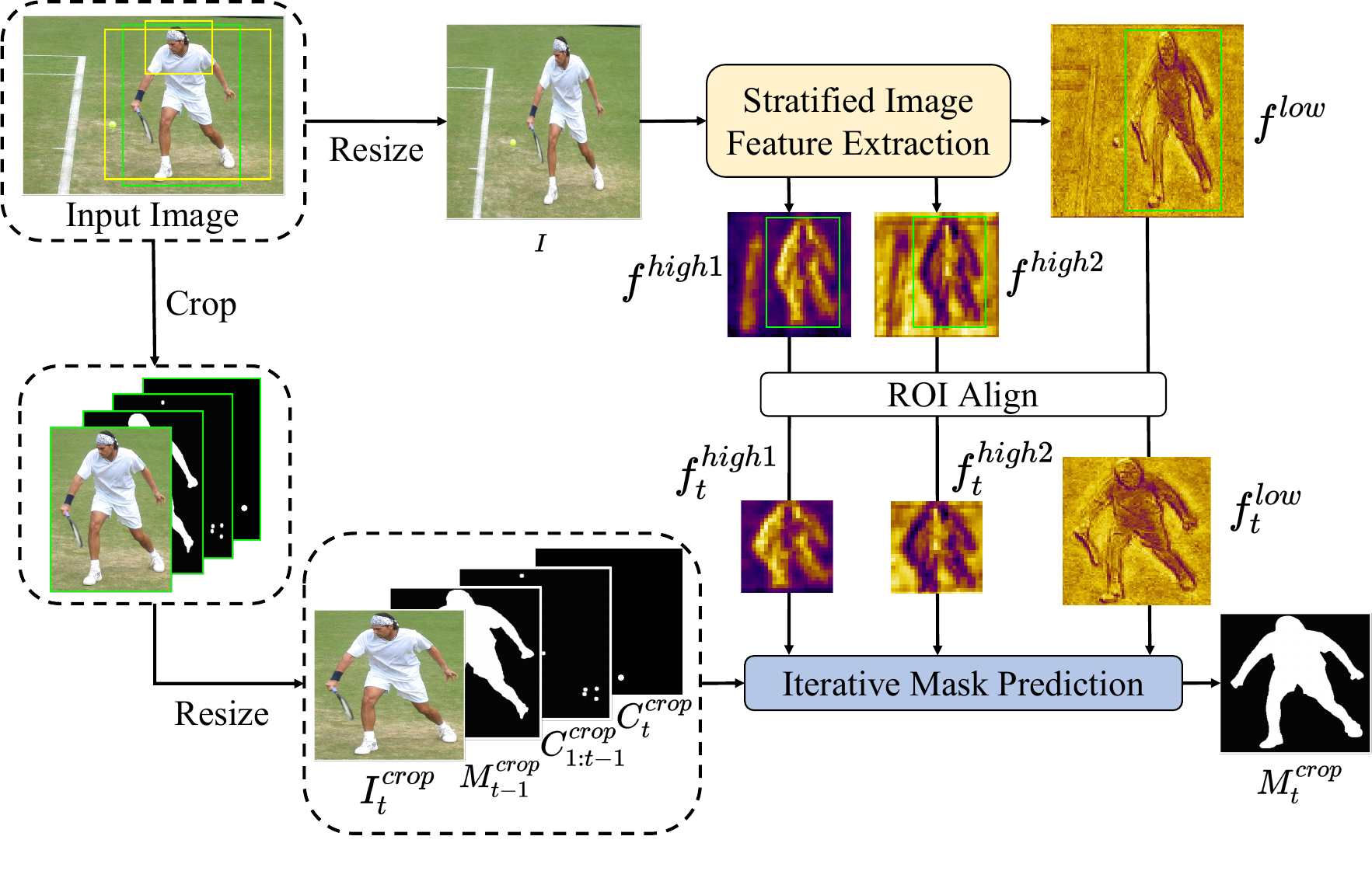}\vspace{-16pt}
    \caption{Visualization of the dynamic-scale training strategy, where yellow bounding boxes denote possible regions of interest and the green one denotes the selected ROI for training.}\vspace{-12pt}
    \label{fig: Dynamic-scale Strategy}
\end{figure}

\subsection{Dynamic-Scale Supervised Learning}\label{Dynamic-scale Strategy}
\textbf{Dynamic-scale training strategy.  } Since the cropped region $I_{t}^{crop}$ varies with each interaction $t$ while the pre-extracted image features are global and invariant, dealing $I$ and $I_t^{crop}$ of different scales and proportions poses challenges to the robustness of our network.  To this end, we propose a simple yet effective dynamic-scale training strategy.   As shown in Fig.~\ref{fig: Dynamic-scale Strategy}, this strategy first generates a region of interest (ROI)  containing the target object with a random proportion of the input image $I$.
We then crop out the local image $I_t^{crop}$ and other inputs (\emph{e.g.,} $M_{t-1}^{crop}$) according to the ROI.    The global input image $I$ and local inputs are separately resized to preset sizes before being fed into the network.  By utilizing this strategy, we ensure that our network can handle diverse combinations of the global $I$ and local $I_t^{crop}$, enabling it to be robust and effective in practical scenarios.  Ablation experiments in Sec.~\ref{sec:Ablation Studies} show that our dynamic-scale training strategy is crucial in facilitating the compatibility between our image features extraction and mask prediction stages.

\noindent \textbf{Loss function. }
We equip two typical interactive segmentation baselines with our decoupling-recycling network: RITM~\cite{sofiiuk2021reviving} and FocalClick~\cite{focalclick}.  The RITM-based models are supervised by the normalized focal loss~\cite{nflloss} $\mathcal{L}_{NFL}$.
For the FocalClick-based models, the loss function  is expressed as follows,
\vspace{-4pt}\begin{equation}
\mathcal{L}=\mathcal{L}_{BCE}+\mathcal{L}_{NFL}+\mathcal{L}_{BNFL},
\end{equation}\vspace{-2pt}
where $\mathcal{L}_{BCE}$, $\mathcal{L}_{NFL}$ and $\mathcal{L}_{BNFL}$ denote the binary cross-entropy loss, the normalized focal loss~\cite{nflloss}  and the boundary normalized focal loss $\mathcal{L}_{BNFL}$~\cite{focalclick}, respectively. The whole model including two stages is trained in an end-to-end manner.

\vspace{-2pt}\begin{table*}[t] 
    \small
    \centering
  \resizebox{0.96\linewidth}{!}{
    \begin{tabular}{l|p{0.76cm}<{\centering}|p{0.74cm}<{\centering}|p{2.4cm}<{\centering}|cc|c|cc|cc}
      \toprule
      \makecell{\multirow{2}{*}{Approaches}} &  \multirow{2}{*}{\makecell{Params\\(M)}} &  \multirow{2}{*}{\makecell{FLOPs\\(G)}} & \multirow{2}{*}{Training Data} &  \multicolumn{2}{c|}{GrabCut~\cite{rother2004grabcut}} &  Berkeley~\cite{martin2001database} &  \multicolumn{2}{c|}{SBD~\cite{sbddataset}} & \multicolumn{2}{c}{DAVIS~\cite{perazzi2016benchmark}} \\
      \cmidrule{5-11}
       &   & &   & \small{ NoC@85} & \small{ NoC@90 }  &  \small{ NoC@90}   &        \small{ NoC@85} & \small{ NoC@90}       &      \small{ NoC@85} & \small{ NoC@90}   \\
     \midrule
     Graph Cut~\cite{boykov2001interactive} & / &  /&  / & 7.98 &  10.00 &  14.22   &  13.60  & 15.96 & 15.13   & 17.41 \\ 
    Geodesic Matting~\cite{gulshan2010geodesic}  &/  & / &  / & 13.32 & 14.57  & 15.96 &  15.36  &  17.60  & 18.59  & 19.50   \\
     Random Walker~\cite{grady2006random}   &/  & / & / & 11.36  & 13.77  &  14.02   &  12.22  & 15.04 & 16.71  &18.31   \\
    Euclidean star~\cite{gulshan2010geodesic} &  / &  /&   / &7.24 &  9.20 &  12.11  & 12.21  &  14.86 & 15.41 &  17.70 \\
    Geodesic star~\cite{gulshan2010geodesic}  & / & / &  / &7.10  &  9.12 & 12.57  &  12.69  & 15.31   & 15.35  & 17.52  \\ 
      \midrule
      Latent~\cite{li2018interactive} & / & / &  SBD~\cite{sbddataset} & 3.20 &  4.79 &  -  &   7.41  &  10.78 &  5.05 &  9.57  \\
    BRS~\cite{jang2019interactive} & /  & / & SBD~\cite{sbddataset} &  2.60  & 3.60  &  5.08   &  6.59  & 9.78  &  5.58 &  8.24 \\
     f-BRS-resnet50~\cite{sofiiuk2020f} &  / &  / & SBD~\cite{sbddataset} & 2.50  & 2.98  &  4.34  & 5.06 &  8.08  & 5.39  &  7.81  \\
    CDNet-resnet50~\cite{chen2021conditional} & 39.90  & 99.16  &  SBD~\cite{sbddataset} &   2.22  &  2.64 &  3.69 &  4.37  &7.87&  5.17 &  6.66  \\
      RITM-hrnet18~\cite{sofiiuk2021reviving}  & 10.03 & 15.07 & SBD~\cite{sbddataset} & 1.76 &  2.04 &   3.22 &   \textbf{3.39} &\textbf{5.43}&  4.94 & 6.71  \\
    Focal-hrnet18s~\cite{focalclick}   & 4.23  & 3.82 & SBD~\cite{sbddataset} &   1.86  & 2.06  & 3.14  & 4.30  & 6.52  & 4.92 &   6.48 \\
    Focal-segformerB0~\cite{focalclick}    & 3.74   & 1.94 & SBD~\cite{sbddataset} &   1.66  &  1.90 & 3.14  & 4.34  &  6.51 & 5.02 &  7.06 \\
    \rowcolor[HTML]{cdcdcd}Ours-H1-hrnet18     &  12.23    & 10.18   & SBD~\cite{sbddataset} &    1.74   &    2.08   &   3.08   &   3.61  &   5.75    &   4.91    & \textbf{6.40} \\
    \rowcolor[HTML]{cdcdcd}Ours-H2-hrnet18s*   &   7.91 &  5.41  &  SBD~\cite{sbddataset} &   1.82  & 1.98  &  \textbf{3.07}  &  4.26   & 6.14   & \textbf{4.72}  &  6.43   \\
    \rowcolor[HTML]{cdcdcd}Ours-B0-segformerB0*    &  7.76  & 2.16   & SBD~\cite{sbddataset} &  \textbf{1.58} & \textbf{1.78} &  3.08  &  4.18  &  6.20  &  4.78  & 6.66  \\
    \midrule
    f-BRS-B-hrnet32~\cite{sofiiuk2020f}   &  /  & / &  COCO~\cite{cocodataset}+LVIS~\cite{lvisdataset} & 1.54   & 1.69   & 2.44  &  4.37 & 7.26  &  5.17 &  6.50 \\
    
    RITM-hrnet18~\cite{sofiiuk2021reviving}     & 10.03 & 15.07 & COCO~\cite{cocodataset}+LVIS~\cite{lvisdataset} &  1.42 &   1.54  & 2.26 &   3.80 &   6.06    &  4.36 &  5.74\\
    RITM-hrnet32~\cite{sofiiuk2021reviving}   & 30.95  & 40.42 & COCO~\cite{cocodataset}+LVIS~\cite{lvisdataset} & 1.46  &1.56  &  2.10   &  3.59 &  5.71 & 4.11 &  5.34  \\
    
    
    Focal-hrnet32~\cite{focalclick}    & 30.98 & 17.12  &   COCO~\cite{cocodataset}+LVIS~\cite{lvisdataset} & 1.64  &   1.80  & 2.36   &  4.24   &   6.51   &  4.01  &  5.39  \\
    Focal-segformerB0~\cite{focalclick}    & 3.74  & 1.94  &  COCO~\cite{cocodataset}+LVIS~\cite{lvisdataset} & \textbf{1.40}    &  1.66  &  2.27  & 4.56 &  6.86  & 4.04 &  5.49 \\
    Focal-segformerB3~\cite{focalclick}   & 45.63 & 12.92  &  COCO~\cite{cocodataset}+LVIS~\cite{lvisdataset} &  1.44  & 1.50  &  1.92 & \textbf{3.53}  & 5.59 & 3.61 & 4.90 \\
    
    \rowcolor[HTML]{cdcdcd}Ours-H2-hrnet18s &  7.90   & 6.65  &  COCO~\cite{cocodataset}+LVIS~\cite{lvisdataset} &   1.46   &  1.66   &   2.23   &  3.79  & 5.85 	& 4.16   &  5.55   \\
    \rowcolor[HTML]{cdcdcd}Ours-H2-hrnet18 & 13.71  &  10.18 & COCO~\cite{cocodataset}+LVIS~\cite{lvisdataset} & 1.46	 &   1.56 &   1.98 &   3.63  &  5.66 &  	3.84	 & 5.32 \\
    
    \rowcolor[HTML]{cdcdcd}Ours-H2-hrnet18s*   & 7.91 &  5.41  &  COCO~\cite{cocodataset}+LVIS~\cite{lvisdataset} &   1.42	 &  1.50	 &  2.07  & 4.19	 & 6.12	 & 3.85 & 5.20 \\
    \rowcolor[HTML]{cdcdcd}Ours-B1-segformerB0* & 13.23 &  2.16 & COCO~\cite{cocodataset}+LVIS~\cite{lvisdataset} & 1.44     &  1.56   & 2.04   &  4.33    &  6.37   &   3.89    & 5.35  \\
    \rowcolor[HTML]{cdcdcd}Ours-B0-segformerB3* & 52.92  & 13.29  &  COCO~\cite{cocodataset}+LVIS~\cite{lvisdataset} &  1.42  &\textbf{1.44}   & \textbf{1.80}   &  3.74&  \textbf{5.57}  &   \textbf{3.55}  & \textbf{4.90}   \\ 
    \bottomrule
    \end{tabular}
    }
    \caption{Quantitative comparisons on GrabCut, Berkeley, SBD and DAVIS dataset, where NoC@85/90 denotes the number of clicks needed to achieve IoU of 85\%/90\%.   Asterisk* indicates models built on FocalClick~\cite{focalclick}. 
    We emphasize the best results in \textbf{bold} and highlight our models with \colorbox[HTML]{cdcdcd}{gray}.  Note that we present FLOPs of the iterative mask prediction stage in this table.
    }\vspace{-22pt}
    \label{performance comparison}
    \end{table*}

 \section{Experiments} \label{sec:Experiments detail}
\subsection{Experimental Settings}\label{sec: Experimental Settings}
\noindent\textbf{Model series. } 
 Although our methods can be applied to almost every backbone, we mainly consider HRNet~\cite{WangSCJDZLMTWLX19} and Segformer~\cite{xie2021segformer} in this paper.  As shown in Tab.~\ref{Speed comparison}, we propose a series of backbone combinations to balance the performance and complexity. For image feature extractor $\mathcal{F}_{\theta 1}$, we choose HRNet-18s, HRNet-18, Segformer-B0 and Segformer-B1, denoted as H1, H2, B0 and B1 respectively, and only keep the first 3 stages of the backbones to reduce unneeded computation complexity. For HRNet~\cite{WangSCJDZLMTWLX19}, the high-level features $f^{high}$ are extracted from the $b^{high}$-th block on the $\frac{1}{16}$-resolution branch, in the $3$rd stage of $\mathcal{F}_{\theta 1}$, and integrate them into the $\widetilde{b}^{high}$-th block in the same branch and stage of $\mathcal{F}_{\theta 2}$. The low-level features $f^{low}$ are extracted from the $b^{low}$-th block on the $\frac{1}{4}$-resolution branch, in the $3$rd stage of the network $\mathcal{F}_{\theta 1}$, and integrate them into the $\widetilde{b}^{low}$-th block in the same branch and stage of the $\mathcal{F}_{\theta 2}$. The values of $b^{high}$, $b^{low} $, $\widetilde{b}^{high}$ and $\widetilde{b}^{low} $ depend on specific network and we empirically set $b^{high} >= b^{low}$, $\widetilde{b}^{high} >= \widetilde{b}^{low}$ to guarantee the high-level features are extracted from deeper layers. We further discuss the choice of these values in Sec.~\ref{sec:Ablation Studies}. Since Segformer~\cite{xie2021segformer} has an encoder-decoder structure, the high-level and low-level features are extracted from the $1$st and $3$rd stage, respectively. We further base our models on two recent interactive segmentation frameworks~\cite{sofiiuk2021reviving,focalclick} with similar pre/post-processing strategies, which we believe are representative and fair.

\noindent\textbf{Datasets.} We conduct evaluations on 4 well-recognized benchmarks: 1) GrabCut~\cite{rother2004grabcut} dataset contains 50 images, as well as a single object mask for each image. 2) Berkeley~\cite{martin2001database} dataset contains 96 images and 100 object masks. 3) SBD~\cite{sbddataset} dataset provides 8,498 images for training and 2,857 images for validation. 4) DAVIS~\cite{perazzi2016benchmark} dataset consists of 345 frames selected from 50 videos. Following~\cite{sofiiuk2020f,sofiiuk2021reviving,focalclick}, we train our models on the combination of COCO~\cite{cocodataset} and LVIS datasets~\cite{lvisdataset}. We also report the performance of our models trained on SBD~\cite{sbddataset} dataset.
 
\noindent\textbf{Evaluation metrics. }  
To quantitatively evaluate the performance, we adopt NoC@IoU as the evaluation metric, which indicates the average number of clicks (NoC) needed to achieve the target intersection over union (IoU) between segmentation results and ground truth masks. A smaller NoC@IoU value indicates better performance since it implies that the target IoU can be achieved with fewer clicks. By convention~\cite{li2018interactive,sofiiuk2021reviving,xu2016deep,focalclick}, the number of clicks is limited to 20, and we simulate clicks by placing the next click at the geometric center of the largest mislabeled region following previous works (\emph{i.e.,} the first click is placed at the center of the ground truth mask). The click simulation will be stopped once the target IoU or the maximum number of clicks is achieved.  

 \begin{figure*}[t]
\centering
\includegraphics[width=0.98\linewidth]{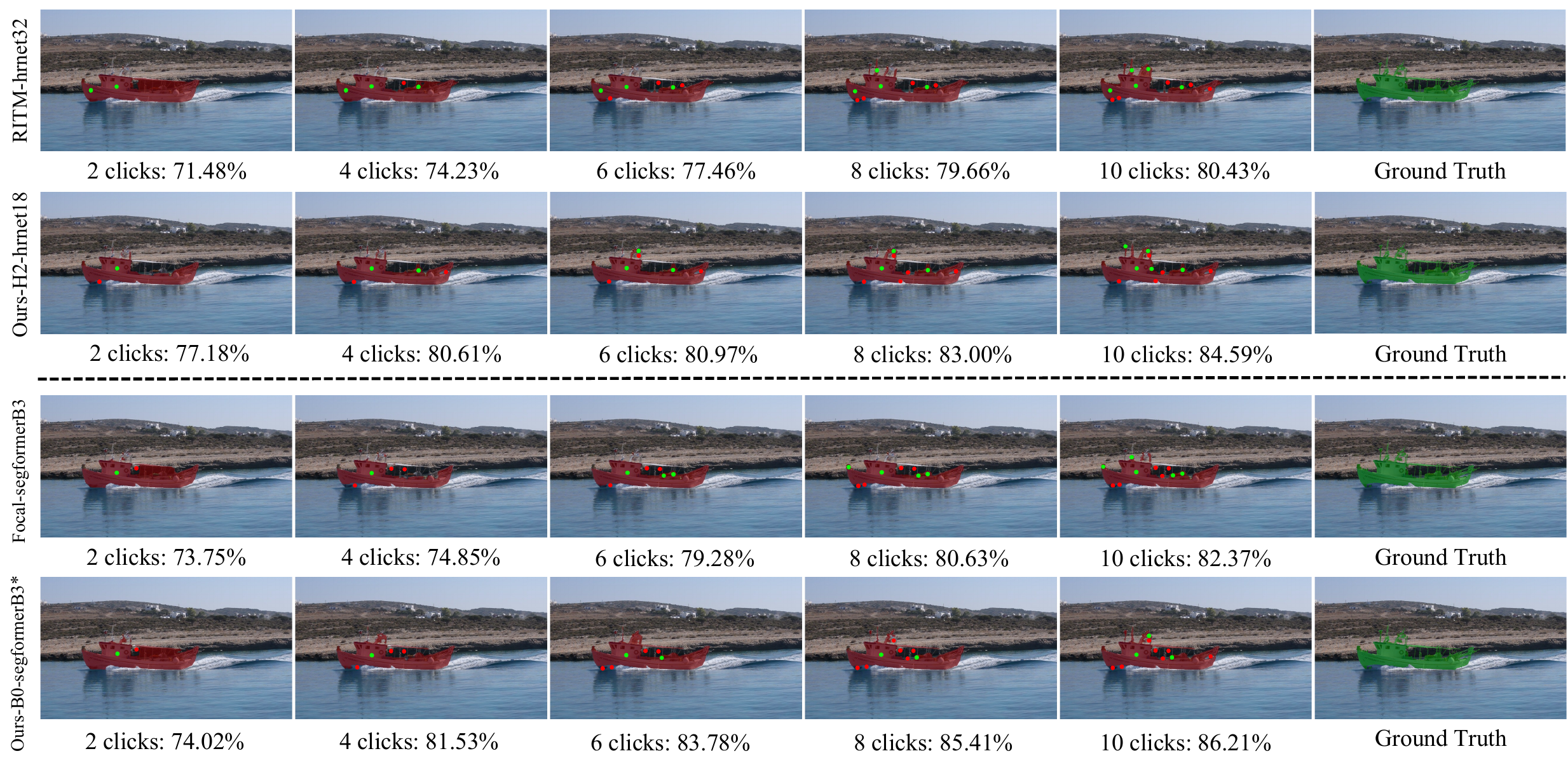}\vspace{-18pt}
\caption{Qualitative comparisons on DAVIS~\cite{perazzi2016benchmark} dataset, where green dots denote positive clicks, and red dots denote negative clicks. We show results with numbers of clicks varying from 2 to 10. Asterisk* indicates that our model is built on FocalClick~\cite{focalclick}.}\vspace{-4pt}
\label{Fig: quali compare}
\end{figure*}

\subsection{Comparisons against Previous Works}\label{sec:Comparison with Previous Works}
In this section, we thoroughly compare our methods with previous works from the following aspects: performance on four benchmarks (\emph{i.e.,} GrabCut, Berkeley, SBD and DAVIS dataset), model complexity and running time. We further discuss the advantages of our methods under long-term clicks situations and evaluate the robustness against misleading user guidance.

\noindent\textbf{Performance comparison. }
The quantitative results on four benchmarks are reported in Tab.~\ref{performance comparison}. According to the training data, we make comparisons in different blocks.  For each method, we report the parameters of the whole model and FLOPs of the iterative mask prediction stage for an intuitive comparison.   As shown in Tab.~\ref{performance comparison}, with similar or slightly more computation cost, our networks outperform previous methods by a large margin (\emph{e.g.,} Ours-B0-segformerB0* \emph{v.s.}  Focal-segformerB0 trained with SBD dataset, and Ours-B0-segformerB3* \emph{v.s.}  Focal-segformerB3 trained with COCO+LVIS dataset).  On the condition of providing better or at least comparable results than existing methods, our method significantly reduces the FLOPs and achieves speedup (\emph{e.g.,} Ours-H1-hrnet18 \emph{v.s.} RITM-hrnet18 trained with SBD dataset, Ours-H2-hrnet18s \emph{v.s.}  RITM-hrnet18, and Ours-H2-hrnet18 \emph{v.s.}  RITM-hrnet32 trained with COCO+LVIS dataset), which is illustrated in the next paragraph in detail. We provide qualitative comparisons in Fig.~\ref{Fig: quali compare}. As can be seen, our method predicts more accurate results and fast converges to high-quality predictions with fewer clicks.

\begin{table}[t]
  \centering
  \resizebox{1\linewidth}{!}{
  \begin{tabular}{l |c |c|c|c|c|c|c|c}
  \toprule
  \makecell{\multirow{2}{*}{Approaches}} &  \multicolumn{2}{c|}{Params/M} &
   \multicolumn{2}{c|}{FLOPs/G} & \multicolumn{2}{c|}{CPU/ms} &  \multicolumn{2}{c}{GPU/ms} \\
  \cmidrule{2-9}
    &$\mathcal{F}_{\theta 1}$  & $\mathcal{F}_{\theta2}$ &
    $\mathcal{F}_{\theta 1}$ & 
    $\mathcal{F}_{\theta2}$ &
    $\mathcal{F}_{\theta 1}$ & 
    $\mathcal{F}_{\theta2}$ &
    $\mathcal{F}_{\theta 1}$  & 
    $\mathcal{F}_{\theta2}$ \\
  \midrule 
  RITM-hrnet18~\cite{sofiiuk2021reviving} & /	  &  10.03  & 	/	 & 15.07  & 	/ &  670 &  / & 33  \\
  RITM-hrnet32~\cite{sofiiuk2021reviving} & /	  & 30.95   & 	/	 & 40.42  & 	/ & 1062 &  / & 50 \\
  \rowcolor[HTML]{cdcdcd}Ours-H2-hrnet18s  & 	
  2.93  &   4.97 &
  7.56  & 6.65 & 	311 &  178	 & 		
  16  &  18 \\
  
  \rowcolor[HTML]{cdcdcd}Ours-H1-hrnet18	 & 	
  1.45 & 10.78 &
  4.11 & 10.18 & 	182  & 234  & 	
 9 & 33   \\
  
  \rowcolor[HTML]{cdcdcd}Ours-H2-hrnet18	 & 		
  2.93  & 10.78 & 
  7.56 & 10.18 & 	
  311 &  234 	 & 		
  16  &  33  \\
  \midrule
  Focal-hrnet18s~\cite{focalclick} & /	  &  4.23  & 	/	 &  3.82 & 	/ & 148  &  / &  17  \\
  Focal-hrnet32~\cite{focalclick} & /	  &   30.98  & 	/	 &  17.12 & 	/ &   288   &  / &  32  \\
  Focal-segformerB0~\cite{focalclick} & /	  & 3.74   & 	/	 &  1.94 & 	/ &   117  &  / & 10 \\
  Focal-segformerB3~\cite{focalclick} & /	  &  45.63  & 	/	 & 12.92  & 	/ &  350  &  / &  24   \\
  \rowcolor[HTML]{cdcdcd}Ours-H2-hrnet18s* & 2.93  & 4.98   &  7.56 &  5.41 &  311  	&  158    & 16 	&  19  \\
  \rowcolor[HTML]{cdcdcd}Ours-B0-segformerB0* & 2.10 &  5.66 &   1.17 &   2.16 & 123  & 	130     & 9  & 10  \\
  \rowcolor[HTML]{cdcdcd}Ours-B1-segformerB0* & 7.57  &  5.66   &  3.82  &  2.16 &  238  & 	130     & 14   & 10  \\
  \rowcolor[HTML]{cdcdcd}Ours-B0-segformerB3* & 2.10  &   50.82	  &  1.17  &  13.29 &  123  & 	 376    & 9  &  25 \\
  \bottomrule
  \end{tabular}} 
  \caption{Comparisons of model complexity and running time, where asterisk* indicate models built on FocalClick~\cite{focalclick}.  Our models are highlighted with \colorbox[HTML]{cdcdcd}{gray}. Note that for each input image, $\mathcal{F}_{\theta 1}$ performs feature extraction only once.}\vspace{-28pt}
  \label{Speed comparison}
  \end{table}

\noindent\textbf{Complexity and running time analysis. } To demonstrate the efficiency of our methods, we report details of model complexity and inference time in Tab.~\ref{Speed comparison}, which directly relates to the user experience.   Since the feature extractor $\mathcal{F}_{\theta 1}$ processes the source input image with the original resolution, we report the FLOPs and inference time of  $\mathcal{F}_{\theta 1}$ with a typical input size of 481$\times$321.   Benefiting from our decoupling-recycling design, the mask prediction network $\mathcal{F}_{\theta 2}$ in our RITM-based models can adopt a smaller input size: 288 $\times$ 288 instead of the  400 $\times$ 400 adopted in RITM~\cite{sofiiuk2021reviving}. For the FocalClick-based models, we keep the 256 $\times$ 256 adopted in FocalClick~\cite{focalclick} as the input size of $\mathcal{F}_{\theta 2}$.    As shown in Tab.~\ref{Speed comparison}, we observe that on the condition of providing comparable or even better results, our models significantly reduce the FLOPs and shorten the inference time.   For instance, RITM-hrnet32 trained on COCO+LVIS takes 5.71 clicks to achieve an IoU of 90\% on SBD dataset (shown in Tab.~\ref{performance comparison}), which costs a total running time of 6,064 ms on a CPU, computed with the following expression,
\vspace{-4pt}\begin{equation}\label{eq: time total}
  t_{total} = t_{\mathcal{F}_{\theta1}} + t_{\mathcal{F}_{\theta2}}*n_{click},
\end{equation}
where $n_{click}$ denote the number of clicks.  $t_{total}$, $t_{\mathcal{F}_{\theta1}}$ and $t_{\mathcal{F}_{\theta2}}$ denote the total running time, time cost of $\mathcal{F}_{\theta1}$ during each interaction and time cost of $\mathcal{F}_{\theta2}$ during each interaction, respectively. In contrast, Ours-H2-hrnet18 takes 5.66 clicks and costs 1,635 ms in total, which is 3.7 times faster.     Compared with Focal-hrnet32 (6.51 clicks/1,875 ms), Ours-H2-hrnet18s* (6.12 clicks/1,278 ms) is 1.47 times faster to achieve an to achieve an IoU of 90\% on SBD dataset. The advantages of our method become more obvious when dealing with more user interactions. For instance, in the case of 20 clicks, the $t_{total}$ of Ours-H2-hrnet18 (4,991 ms) is 4.25 times less than that of RITM-hrnet32 (21,240 ms).

\begin{figure}[t]
  \centering
 \vspace{-2pt}\includegraphics[width=0.98\linewidth]{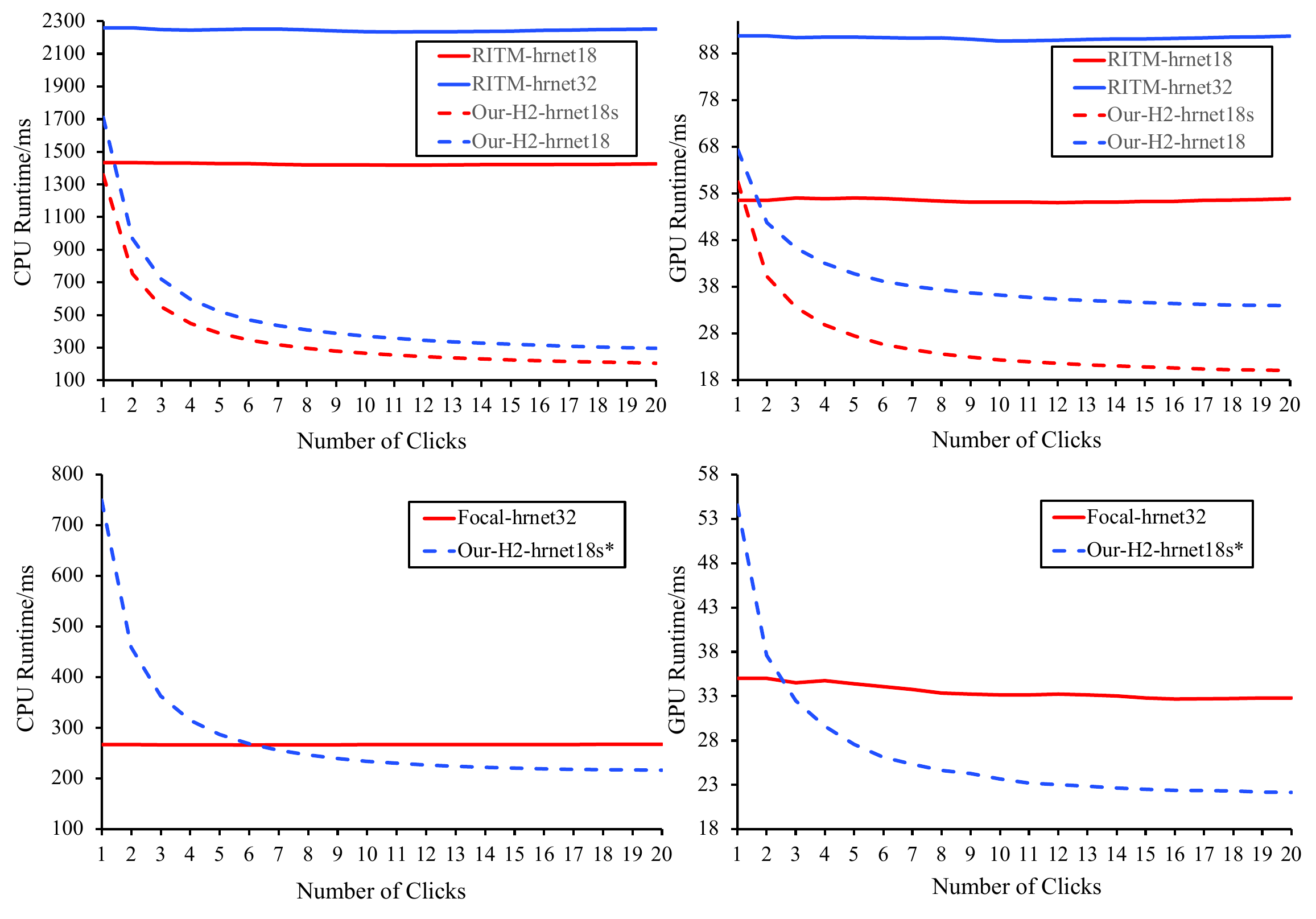}\vspace{-12pt}
  \caption{Average time cost of each interaction in the situation of long-term clicks.  We test the average time cost of each interaction on 50 images from DAVIS~\cite{perazzi2016benchmark} dataset and set the maximum number of clicks to 20.}\vspace{-12pt} 
  \label{fig: long term clicks}
\end{figure}

\noindent\textbf{Long-term clicks time cost. }  
For objects with complex structures and detailed edges, high-quality segmentation requires amounts of user inputs. Applying our framework to such a situation significantly improves efficiency. We select 50 images from the DAVIS dataset which require more than 20 clicks for existing methods to achieve an IoU of 90\%.  Setting the maximum number of clicks to 20, we compute the average time cost of each interaction for representative models with the expression as follows,
\vspace{-4pt}\begin{equation}\label{eq: long term}
  t_{ave} = t_{\mathcal{F}_{\theta1}}/n_{click} + t_{\mathcal{F}_{\theta2}},  
\end{equation}
where $n_{click}$ denote the number of clicks. $t_{ave}$, $t_{\mathcal{F}_{\theta1}}$ and $t_{\mathcal{F}_{\theta2}}$ denote the average time cost, time cost of $\mathcal{F}_{\theta1}$ and time cost of $\mathcal{F}_{\theta2}$ during each interaction, respectively. As shown in Fig.~\ref{fig: long term clicks}, compared with RITM~\cite{sofiiuk2021reviving} and FocalClick~\cite{focalclick}, the average time cost $t_{ave}$ of our method reduces with the number of input clicks $n_{click}$, while $t_{ave}$ of the other methods stays at constant. And the average inference time infinitely approaches the time cost of the adopted mask prediction network if more clicks are included. Such a trend also indicates that our framework can be applied to segmenting multi-objects within a single input image to promote efficiency.

\begin{table}[t]
\begin{center}
  \resizebox{0.9\linewidth}{!}{
  \begin{tabular}{l|cc|cc}
  \toprule
  \makecell{\multirow{2}{*}{Models}} &   \multicolumn{2}{c|}{SBD~\cite{sbddataset}} & \multicolumn{2}{c}{DAVIS~\cite{perazzi2016benchmark}} \\
  \cmidrule{2-5}
  & NoC@85 & NoC@90 &NoC@85 & NoC@90 \\
  \midrule
  RITM-hrnet18~\cite{sofiiuk2021reviving}  & 5.07   &  7.68&  5.13   &  6.91 \\
  RITM-hrnet32~\cite{sofiiuk2021reviving}  &   4.83  &  7.36  & 4.92   &   6.18  \\
  Focal-hrnet32~\cite{focalclick}  &5.70  &   8.34 & 4.61    &  6.15   \\
  Focal-segformerB3~\cite{focalclick}  & 4.91   &  7.31  & 4.15    &  \textbf{5.49}  \\
  Ours-H2-hrnet18s  & 4.83   &  7.27  & 4.71   &   6.34 \\
  Ours-H2-hrnet18  &\textbf{4.63}  &  \textbf{6.97}  & 4.70   &   6.08  \\
  Ours-H2-hrnet18s*  &  5.64   &  7.95 & 4.40   &   5.86 \\
  Ours-B0-segformerB3*  & 5.11   &  7.29 & \textbf{4.12}    &  5.66 \\
  \bottomrule
  \end{tabular}}
  \end{center}
   \caption{Robustness evaluation in realistic scenarios with repetitive click and false click.  Our method is more robust than other methods.}\vspace{-26pt}
  \label{tab: user intents}
  \end{table}

\noindent\textbf{Robustness evaluation on misleading user guidance. }
Existing methods ~\cite{sofiiuk2021reviving,jang2019interactive,sofiiuk2020f}  iteratively place new clicks at the center of the largest mislabeled region and assume user guidance is always correct.   To simulate more realistic user guidance, we consider the following situations that challenge the robustness of interactive segmentation models: 1) repetitive click, which generates the next click within the previously predicted mask region;   2) false click,  which indicates the click type (positive/negative) is given by mistake. For each test sample, we first randomly define 5 clicks out of 20 clicks to be repetitive or false clicks, while defining the rest 15 clicks as effective interactions.  Then we apply the same sequence of click definitions to all models.  Results of the SBD~\cite{sbddataset} and DAVIS~\cite{perazzi2016benchmark} dataset are shown in Tab.  ~\ref{tab: user intents}.  As can be seen, confronting practical scenarios above, the performance of previous methods drops significantly while our models still perform beyond the others, which proves the robustness of our model.  We attribute this to the decoupling structure of our network.   Firstly, by decoupling the current and historical clicks, we highlight the effect of the current click so that mislabeled areas introduced by previous repetitive/false clicks can be effectively corrected;   Secondly, the decoupled high-level features helps to propagate the correct user clicks to long-range areas.   Both designs can quickly correct misleading user guidance.

\begin{table}[t]
\begin{center}
  \resizebox{0.9\linewidth}{!}{
  \begin{tabular}{l|cc|cc}
  \toprule
  \makecell{\multirow{2}{*}{Models}} &   \multicolumn{2}{c|}{GlaS~\cite{glas}} & \multicolumn{2}{c}{BraTS~\cite{brats}} \\
  \cmidrule{2-5}
  & NoC@85 & NoC@90 &NoC@85 & NoC@90 \\
  \midrule
  RITM-hrnet18~\cite{sofiiuk2021reviving}  &  5.64    & 7.28   &    9.67 &  11.46  \\
  RITM-hrnet32~\cite{sofiiuk2021reviving}  &   5.47   &  6.94    &    8.46    &  11.88   \\
  Focal-hrnet32~\cite{focalclick}  &   6.21    &  8.34    &     9.32    &  10.59   \\
  Focal-segformerB3~\cite{focalclick} &   4.04   &  5.36    &   8.51    &   10.10   \\
  Ours-H2-hrnet18s & 5.59  &  7.26  &    9.45  &   11.20     \\
  Ours-H2-hrnet18 &   5.24   &  6.92    &  \textbf{7.90}   &  9.76      \\
  Ours-H2-hrnet18s*  &  4.63   &   6.38    &   9.07   &  10.38      \\
  Ours-B0-segformerB3* &    \textbf{3.55}  & \textbf{4.94}   &     8.50   &  \textbf{9.72}    \\
  \bottomrule
  \end{tabular}
  }
\end{center}
\caption{Quantitative comparisons on medical segmentation dataset GlaS~\cite{glas} and BraTS~\cite{brats}, where evaluation on GlaS~\cite{glas} dataset is conducted without fine-tuning and we fine-tune all models on BraTS~\cite{brats} dataset for 30 epochs.}\vspace{-20pt}
\label{tab: medical}
\end{table}

\vspace{-8pt}\subsection{Generalization to Medical Segmentation}
\noindent \textbf{Datasets and experiment settings.} In this section, to assess the generalization ability of our methods, we conduct cross-domain evaluations on two medical datasets:  1) GlaS~\cite{glas} dataset consists of 80 microscopy images of glands in colon histology and provides annotations for each gland.  2) BraTS~\cite{brats} dataset consists of 1,251 brain tumor volumes in MRI scans for training and 2 volumes for evaluation, and provides ground truth masks for every tumor sub-region. 
These medical datasets distinctly differ from natural images in terms of data domains and modalities. Specifically, we apply models trained on COCO+ LVIS dataset to GlaS~\cite{glas} dataset without fine-tuning.  For BraTS~\cite{brats} dataset, we randomly select 500 volumes from the training split and fine-tune all models for 30 epochs.  Gray-scale images are channel-wisely repeated 3 times to meet the requirement for input size. To be fair, we apply the same click generation strategy mentioned in Sec.~\ref{sec: Experimental Settings} to all methods.

\noindent\textbf{Evaluation of cross-domain generalization.} 
We report quantitative results of GlaS~\cite{glas} and  BraTS~\cite{brats} dataset in Tab.~\ref{tab: medical}.  As can be seen, our models show favorable performance and better generalization ability than existing methods. For instance, Focal-hrnet32 requires 8.34 clicks to achieve an IoU of 90\% on the GlaS~\cite{glas} dataset, while Ours-H2-hrnet18s* only needs 6.38 clicks. Ours-H2-hrnet18s* achieves IoU of 90\% with 2 fewer clicks than Focal-hrnet32 on the GlaS~\cite{glas}  dataset, and Ours-H2-hrnet18 achieves IoU of 90\% with 2 fewer clicks than RITM-hrnet32 on the  BraTS~\cite{brats} dataset.

\begin{figure}[t]
\hspace{-10pt}\includegraphics[width=0.925\linewidth]{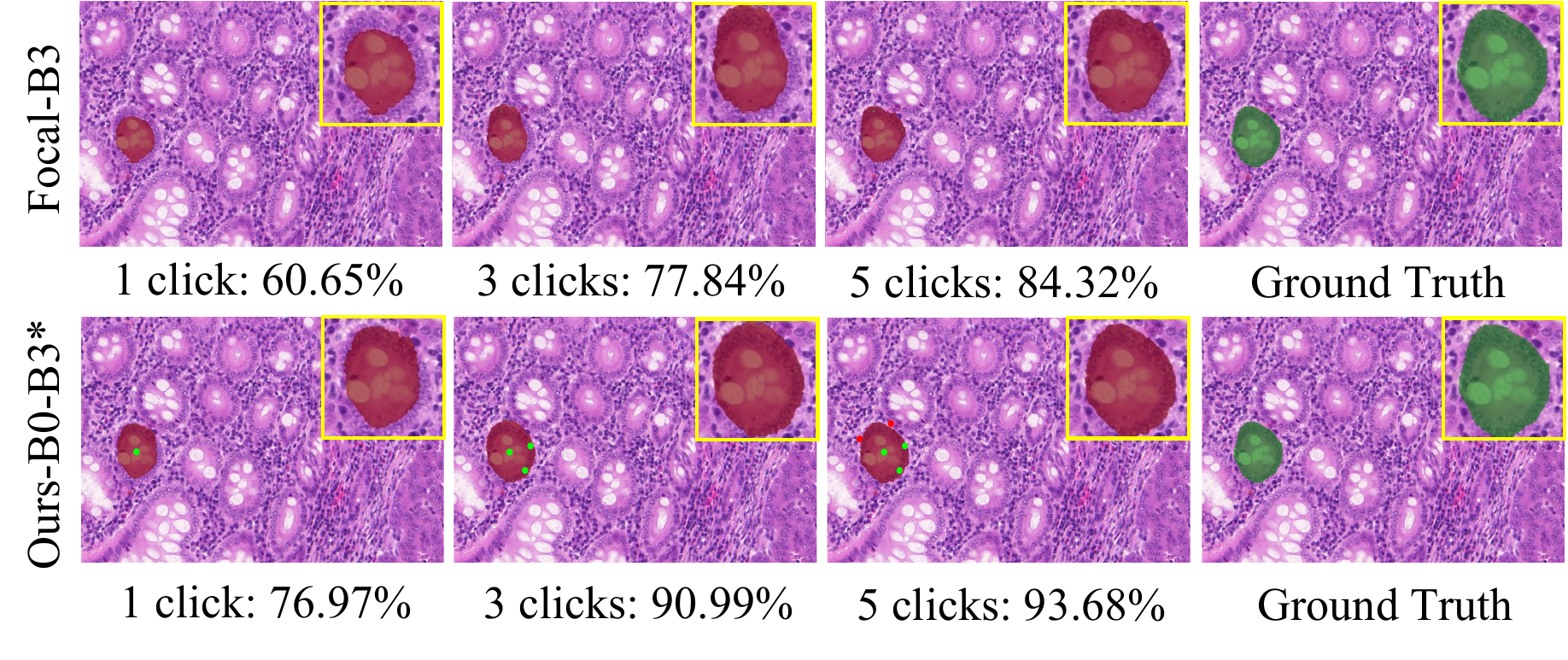}\\
\vspace{-4pt} \hspace{-11pt}\includegraphics[width=0.925\linewidth]{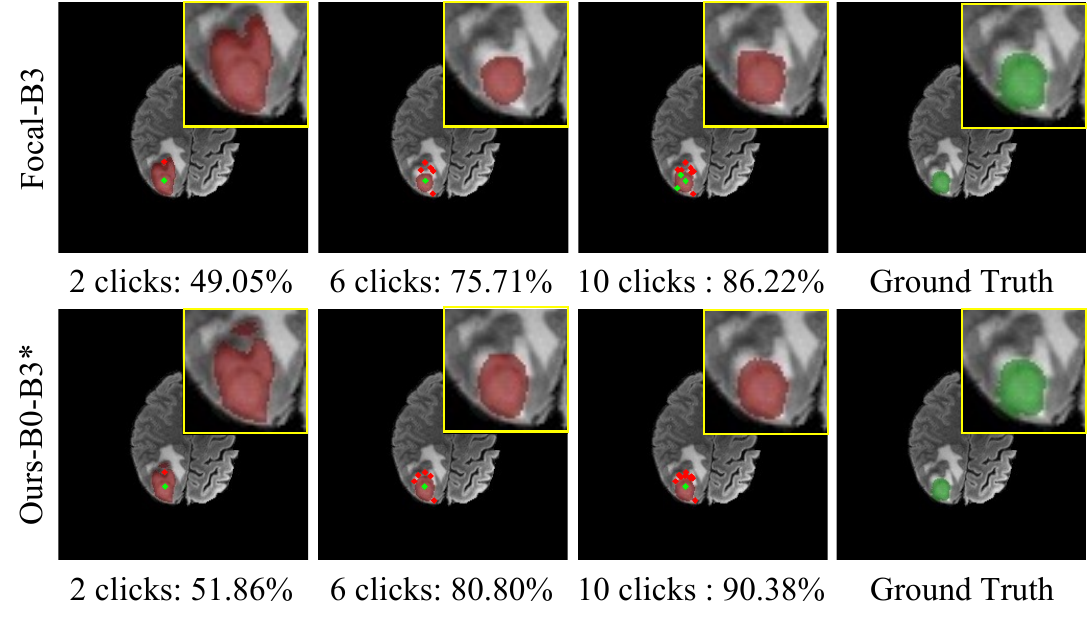}\vspace{-10pt}
\caption{Qualitative comparisons on medical segmentation dataset GlaS~\cite{glas} and BraTS~\cite{brats}, where green/red dots denote positive/negative clicks. `B3' is short for segformerB3.}\vspace{-12pt}
\label{fig: medical}
\end{figure}

\noindent\textbf{Challenging task requiring long-term user interactions.}
We provide qualitative segmentation comparisons in Fig.~\ref{fig: medical}, which demonstrate the effectiveness of our methods in providing favorable segmentation results with limited clicks,  and the capability to refine roughly predicted boundaries.
Moreover, since medical images often contain multiple objects to be annotated  (\emph{e.g.,} glands in the 1st case of Fig.~\ref{fig: medical}) and require elaborate annotations for edges (\emph{e.g.,} brain tumor in the 2nd case of Fig.~\ref{fig: medical}), long-term clicks are needed in these challenging cases, and our method provides a practical solution to elevate the efficiency of medical image annotation. 

\vspace{-4pt} 

\subsection{Ablation Studies}\label{sec:Ablation Studies}
We perform ablation studies for the following problems: 1) For image features extraction network $\mathcal{F}_{\theta 1}$, which level of features are essential and should be reused? 2) Where to extract these features from $\mathcal{F}_{\theta 1}$ and integrate them into $\mathcal{F}_{\theta 2}$? 3) The effectiveness of dynamic-scale training strategy. 4) The effectiveness of decoupling current and historical user guidance.  To keep generality, we perform ablation studies on Ours-H2-hrnet18s and report results of DAVIS and SBD dataset.  All models are trained on COCO+LVIS dataset.

 \begin{table}[t]
  \centering
  \resizebox{0.83\linewidth}{!}{
  \begin{tabular}{c|c|c|cc|cc}
  \toprule
    \multirow{2}{*}{$f_{low}$} & \multirow{2}{*}{$f_{mid}$} & \multirow{2}{*}{$f_{high}$} & \multicolumn{2}{c|}{SBD~\cite{sbddataset}} & \multicolumn{2}{c}{DAVIS~\cite{perazzi2016benchmark}}  \\
  \cmidrule{4-7}
   &  &   &NoC@85 & NoC@90 &NoC@85 & NoC@90 \\
  \midrule
    -&- & - &  4.04 &	6.48	& 4.70 &	5.98  \\
  -&- &\checkmark &   3.93   & 6.02   &  4.22  & 5.77    \\
    \checkmark&- & -&  4.00 &  6.04  &   4.25 &  5.81    \\
   \checkmark & - &\checkmark & \textbf{3.80}  &  \textbf{5.83}	&  4.23   & \textbf{5.55} \\
  \checkmark&\checkmark &\checkmark &   3.93   &  5.96    & \textbf{4.13}    & 5.83   \\
  \bottomrule
  \end{tabular}}
   \caption{Ablation studies of extracted features from the image feature extractor $\mathcal{F}_{\theta 1}$.}\vspace{-24pt}
  \label{which feature ablation}
  \end{table}

\noindent\textbf{Effectiveness of extracted image features. } 
As stated in Sec.~\ref{sec: Experimental Settings}, we take HRNet~\cite{WangSCJDZLMTWLX19} without the last stage as the image features extraction network $\mathcal{F}_{\theta 1}$ for a performance-speed balance.
In the $3$rd stage, there are three parallel branches with features at different resolutions (\emph{i.e.,} $\frac{1}{4}$, $\frac{1}{8}$ and $\frac{1}{16}$). We distinguish them by the extracted branch and name as low-level features $f^{low}$, middle-level features $f^{mid}$ and high-level features $f^{high}$. We report the results of recycling features of different levels in Tab.~\ref{which feature ablation}. As can be seen, our method equipped with $f^{high}$ and $f^{low}$ achieves the best performance.

  \noindent\textbf{Positions of image features.   } We discuss different combinations of hyper-parameters $b^{high}$, $b^{low}$, $\widetilde{b}^{high}$, and  $\widetilde{b}^{low}$, indicating different image feature extraction and integration positions in $\mathcal{F}_{\theta 1}$ and $\mathcal{F}_{\theta 2}$.   We seek to find an optimal combination through two steps: 1) We fix the positions of the low-level features $b^{low}$ and $\widetilde{b}^{low}$,  and change the positions of high-level features $b^{high}$ and $ \widetilde{b}^{high}$; 2) Then we select another position for $b^{low}$ and $\widetilde{b}^{low}$, and repeat the operation of 1) until all combinations are included.    By default, we set $\widetilde{b}^{high} = b^{high}$  and $ \widetilde{b}^{low} = b^{low}$ unless the number of multi-resolution blocks in $\mathcal{F}_{\theta 1}$ and $\mathcal{F}_{\theta 2}$ are not the same.   The results are reported in Tab.  ~\ref{tab: feature choose}.   We adopt the combination of $b^{low}$ = 3, $b^{high}$ = 3, $\widetilde{b}^{low}$ = 2  and $\widetilde{b}^{high}$=2  due to its superiority over the others.

\begin{table}[t]
  \centering
 \vspace{-4pt}  \resizebox{0.83\linewidth}{!}{
  \begin{tabular}{c|c|c|c|cc|cc}
  \toprule
  \multirow{2}{*}{$b^{low}$} &\multirow{2}{*}{$b^{high}$} & \multirow{2}{*}{$\widetilde{b}^{low}$} & \multirow{2}{*}{$\widetilde{b}^{high}$} & \multicolumn{2}{c|}{SBD~\cite{sbddataset}} & \multicolumn{2}{c}{DAVIS~\cite{perazzi2016benchmark}}  \\
   \cmidrule{5-8}
  & & &   &NoC@85 & NoC@90 &NoC@85 & NoC@90 \\
  \midrule
  0 & 0 & 0  &   0 &      3.92  &   5.98    & \textbf{4.22}  &  5.67      \\
  0 & 1 & 0  &   1 &   3.84   &  5.90  &  4.25 & 5.61     \\
  0 & 2 & 0  &   2 &  3.82  &   5.86    &  4.30  &   5.74   \\
  0 & 3 & 0  &   2 &    3.83   &   5.87  & 4.32  &  5.59    \\
  \midrule
  1 & 1 & 1  &   1 &   3.90   &  5.97     &   4.32    &  5.65     \\
  1 & 2 & 1  &   2 &   \textbf{3.79}  &  5.86      & 4.30  &  5.56    \\
  1 & 3 & 1  &   2 &   3.82   &  5.96     &  4.34   &   5.63   \\         
  \midrule
  2 & 2 & 2  &   2 &    3.89     & 5.94       &   4.23    &  5.63     \\
  2 & 3 & 2  &   2 &       3.91    & 6.00   & 4.36    &  5.70      \\
  \midrule
  3 & 3 & 2  &   2 &     3.80   &    \textbf{5.83}   &  4.23  &   \textbf{5.55}   \\
  
  \bottomrule
  \end{tabular}
  }
  \caption{Ablation studies on positions of extracting and integrating features, where $b^{low}$/$b^{high}$ indicate positions of extracting features from $\mathcal{F}_{\theta 1}$, $\widetilde{b}^{low}$/$\widetilde{b}^{high}$ indicate positions of integrating features into $\mathcal{F}_{\theta 2}$. }\vspace{-16pt} 
  \label{tab: feature choose}
  \end{table}

  \begin{table}[t]
    \centering

    \resizebox{0.83\linewidth}{!}{
    \begin{tabular}{p{1.5cm}<{\centering}|c|cc|cc}
    \toprule
    \multirow{2}{*}{\shortstack{Historical \\ decoupling}} & \multirow{2}{*}{\shortstack{Dynamic\\strategy}}  & \multicolumn{2}{c|}{SBD~\cite{sbddataset}} & \multicolumn{2}{c}{DAVIS~\cite{perazzi2016benchmark}}   \\
    \cmidrule{3-6}
   & & NoC@85 & NoC@90 &NoC@85 & NoC@90 \\
    \midrule
      \checkmark & - & 3.95  &  6.08    & 4.47   & 5.74   \\
       - &   \checkmark  &   3.80  &  \textbf{5.83}     &   4.23   & 5.55      \\
           \checkmark &  \checkmark &  \textbf{3.79}   &  5.85 	&  \textbf{4.16}   & \textbf{5.55}  \\
    \bottomrule
    \end{tabular}}
    \caption{Ablation studies of the dynamic-scale training strategy and current-historical guidance decoupling.}\vspace{-16pt} 
    \label{tab: decouple and dynamic}
    \end{table}

\noindent\textbf{Effectiveness of dynamic-scale training strategy. }
To verify the effectiveness of the dynamic-scale training strategy, we remove it and train with source image $I$ and local image $I_t^{crop}$ of fixed-proportion.  As illustrated in the 1st row of Tab.~\ref{tab: decouple and dynamic}, the performance on the two benchmarks drops significantly without the dynamic-scale training strategy, demonstrating its effectiveness.

\noindent\textbf{Effectiveness of current-historical guidance decoupling. }  To investigate the effectiveness of decoupling historical and current user guidance, we replace it with simultaneously encoding all clicks~\cite{sofiiuk2021reviving,focalclick}. As illustrated in the 2nd row of Tab.~\ref{tab: decouple and dynamic}, simultaneously encoding all clicks shows sub-optimal results compared with decoupling current and historical clicks.

\section{Conclusion}
In this paper, we propose a simple yet efficient framework for interactive segmentation. Our main idea is to decouple 3 core components and reuse extracted features. With such designs, our method extracts more effective image representations, while achieving significant speedup. Ablation studies prove the effectiveness of our proposed decoupling designs. Experimental results on various datasets also demonstrated the superiority over existing methods. We believe our method can also be easily integrated into other frameworks and benefit future works in this area.

\noindent\textbf{Limitations. }
So far, we have only explored two kinds of backbones and constructed models with variants of the same backbone so that networks of the image feature stage and mask prediction stage share similar structures and compatible features.
We will further explore more combinations of diverse networks in future work.

\begin{acks}
This work was supported in part by the NSFC (U2013210, 62006060), in part by the Guangdong Basic and Applied Basic Research Foundation under Grant (2022A1515010306),  in part by the Shenzhen Fundamental Research Program under Grant (JCYJ20220818102415032).
\end{acks}

\clearpage 
\bibliographystyle{ACM-Reference-Format}
\balance
\bibliography{egbib.bib}

\clearpage
 \nobalance
\appendix
\section{Appendix}

\subsection{Implementation Details}\label{details}

\noindent\textbf{Experimental settings. } During training, the proportion of ROI in the dynamic-scale training strategy is randomly selected between 0.3 and 1. The global image is resized to 480 $\times$ 480. The cropped region is resized to 320 $\times$ 320 and 256 $\times$ 256 in RITM-based models and FocalClick-based models, respectively.  The initial number of positive/negative clicks is randomly generated between 1 and 24, and the maximum number of interactions is randomly generated between 1 and 3. We train our models on the SBD~\cite{sbddataset} dataset and COCO~\cite{cocodataset}+LVIS~\cite{lvisdataset} dataset for 230 and 300 epochs, respectively. The batch size is set to 48. We adopt the Adam optimizer~\cite{adamopti} with $\beta1=0.9$ and $\beta2=0.999$. The learning rate is initialized as 5e-4 and decreases by 10 times at the epoch of 190 and 210. During inference, we follow previous works~\cite{sofiiuk2021reviving,focalclick} to determine and resize $I_t^{crop}$ with the Zoom-In strategy~\cite{sofiiuk2020f}.  Our proposed framework is implemented with PyTorch~\cite{paszke2019pytorch}. The running time is measured on a PC with 2.9 GHz, Core i7 and NVIDIA GeForce GTX 1660Ti. 

\noindent \textbf{Click simulation strategy.}  
Following previous works~\cite{sofiiuk2021reviving,focalclick,sofiiuk2020f}, our user-click simulation strategy generates the next click at the center of the largest mislabeled region. This region is determined by comparing the segmented result with the ground truth mask.  According to the type of the mislabeled region (false negative/false positive), the type of the next click is determined to be positive/negative.  Adopting this strategy for evaluation ensures that the simulated clicks are dynamically generated based on the result of each interaction rather than real user-given clicks, thus preventing potential unfair comparisons.

\noindent \textbf{Details of the Brats dataset. }
Although MRI volumes are provided in the Brats~\cite{brats} dataset, brain tumors do not exist across the whole 3D volume. As a result, some 2D slices are devoid of any brain tumors and the masks are zero. Given that clicks are simulated according to the difference between the current result and the ground truth mask, we remove tumor-free 2D slices and conduct experiments on 2D slices with non-zero ground truth masks.

\subsection{More Visual Results}\label{visualizations}
\begin{figure}[t]
  \centering
  \includegraphics[width=0.955\linewidth]{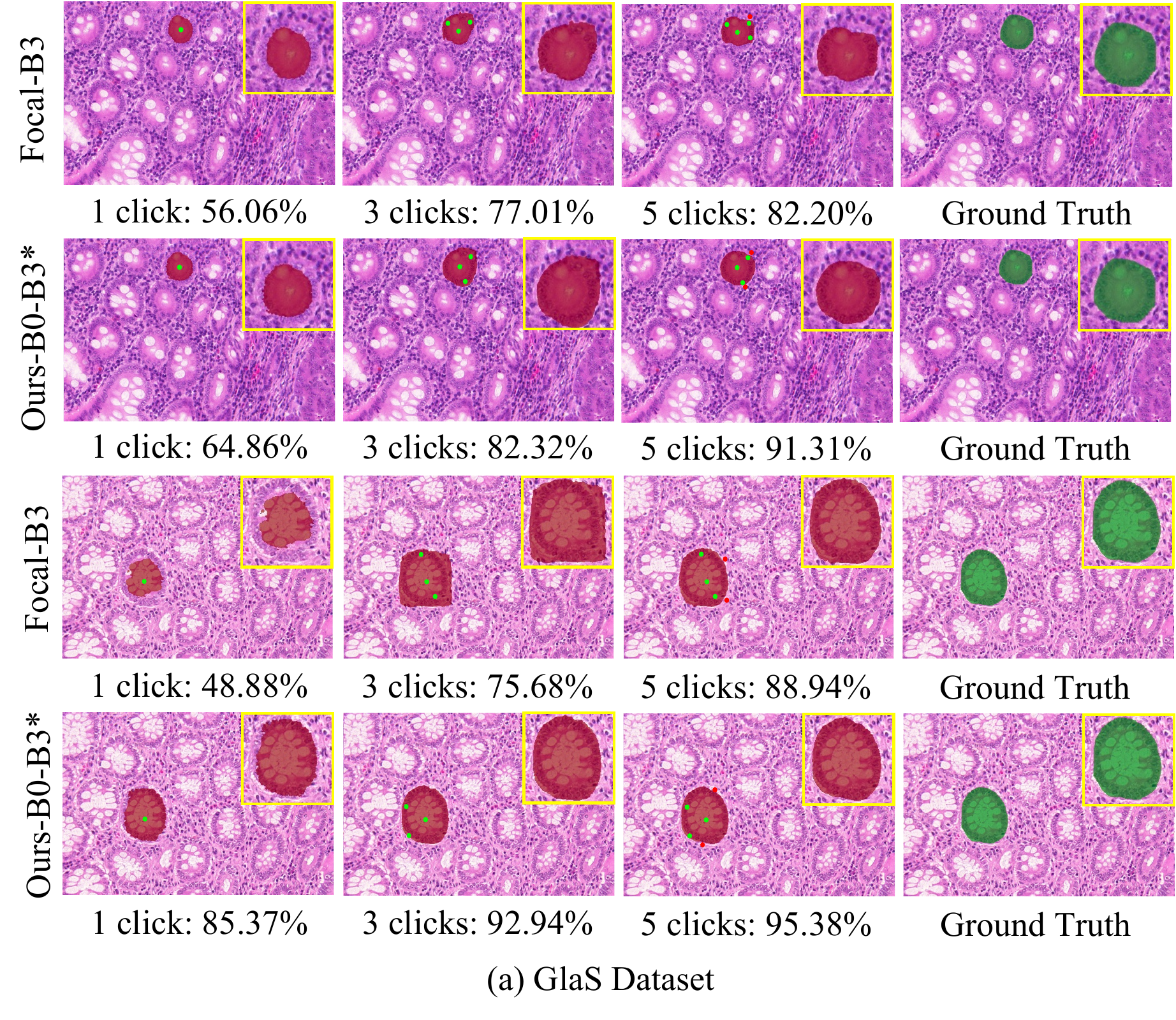}
  \includegraphics[width=0.955\linewidth]{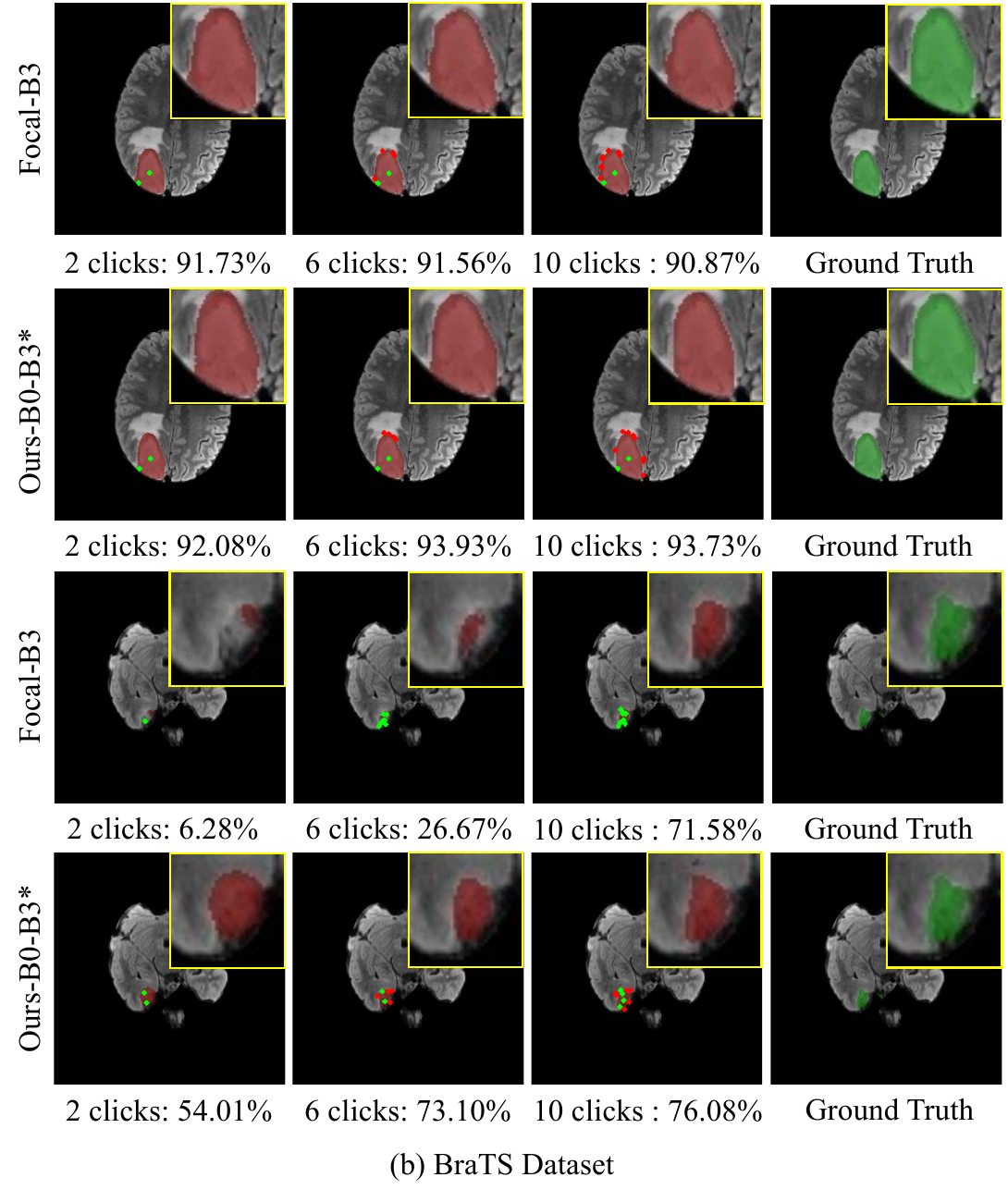}
  \vspace{-13pt}\caption{Qualitative comparisons on GlaS~\cite{glas} dataset and BraTS~\cite{brats} dataset, where green/red dots denote positive/negative clicks. `B3' is short for segformerB3.}\vspace{-10pt} 
  \label{fig_supp: medical}
  \end{figure}

\noindent\textbf{Qualitative comparisons of medical images.} 
We include more qualitative results of medical images in Fig.~\ref{fig_supp: medical}. As shown in Fig.~\ref{fig_supp: medical}(a), our method achieves remarkable visual results with limited interactions, while consistently refining the edges of predicted masks.  As can be seen from Fig.~\ref{fig_supp: medical}(b), our method effectively employs user guidance to annotate objects with irregular shapes. It is worth noting that, annotating multiple objects within a single image  (\emph{e.g.,} glands in Fig.~\ref{fig_supp: medical}(a)) or mask refinement with multiple clicks (\emph{e.g.,} tumors with irregular shapes in Fig.~\ref{fig_supp: medical}(b)) require a considerable amount of user interactions. In these situations, our method shows superior efficiency over existing methods by recycling extracted image semantics.

\begin{figure*}[t] 
  \hspace{-10pt}\includegraphics[width=1\linewidth]{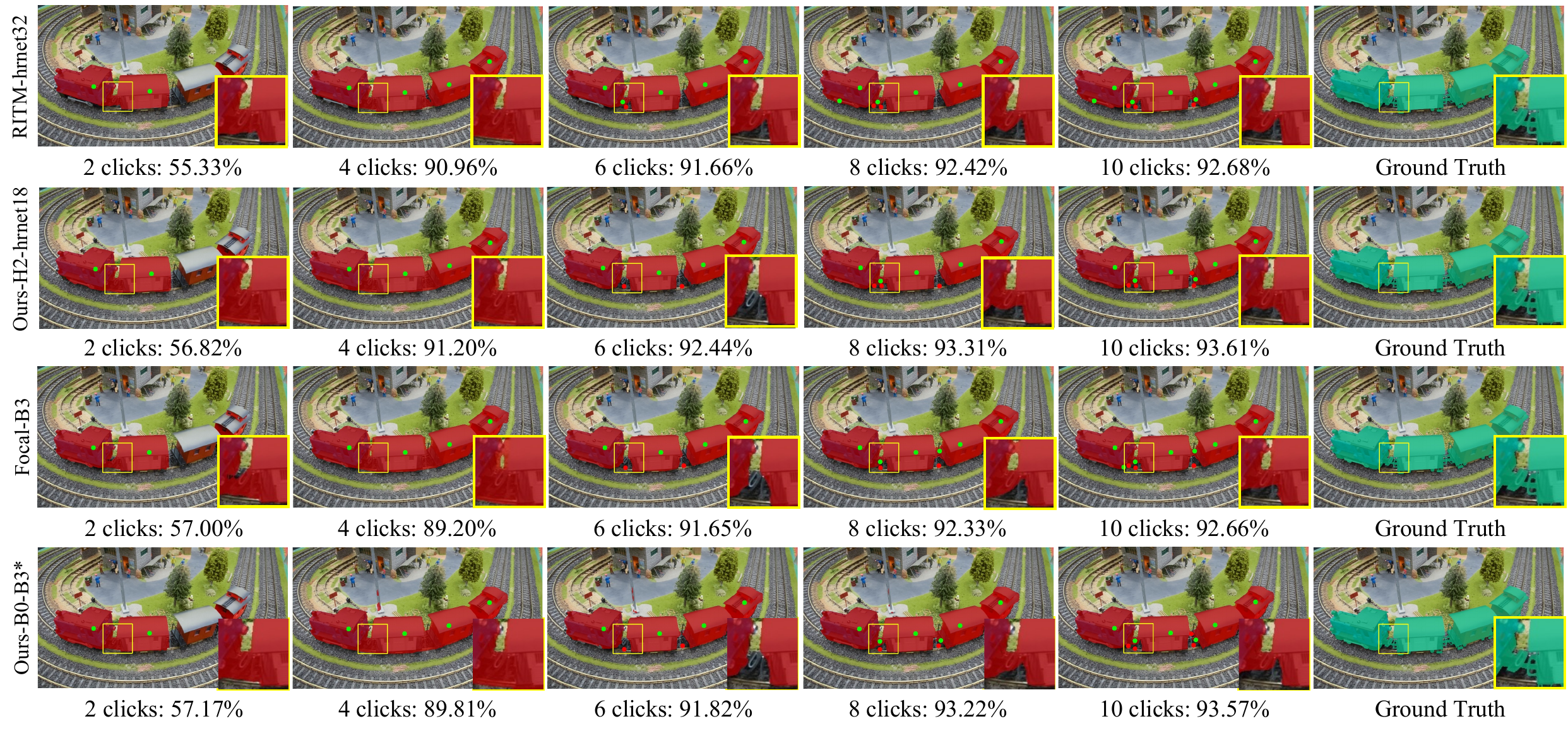}\\
  \vspace{-10pt}\caption{Qualitative comparisons on DAVIS~\cite{perazzi2016benchmark} dataset, where green/red dots denote positive/negative clicks. `B3' is short for segformerB3.}\vspace{-4pt}  
  \label{fig_supp: natural}
  \end{figure*}

\noindent\textbf{Qualitative comparisons of natural images. } 
We provide additional qualitative comparisons of natural images in Fig.~\ref{fig_supp: natural}. As can be seen, our method predicts more accurate results for intricate structures  (\emph{e.g.,} hinges of the train in the 1st case of Fig.~\ref{fig_supp: natural}) and objects that are easily to be mixed with the background (\emph{e.g.,} quill of the swan in the 2nd case of Fig.~\ref{fig_supp: natural}). These results indicate the effectiveness of our approach in utilizing user guidance and image semantics in complicated scenarios.

\begin{figure*}[t] 
\centering
\vspace{-10pt}\includegraphics[width=0.98\linewidth]{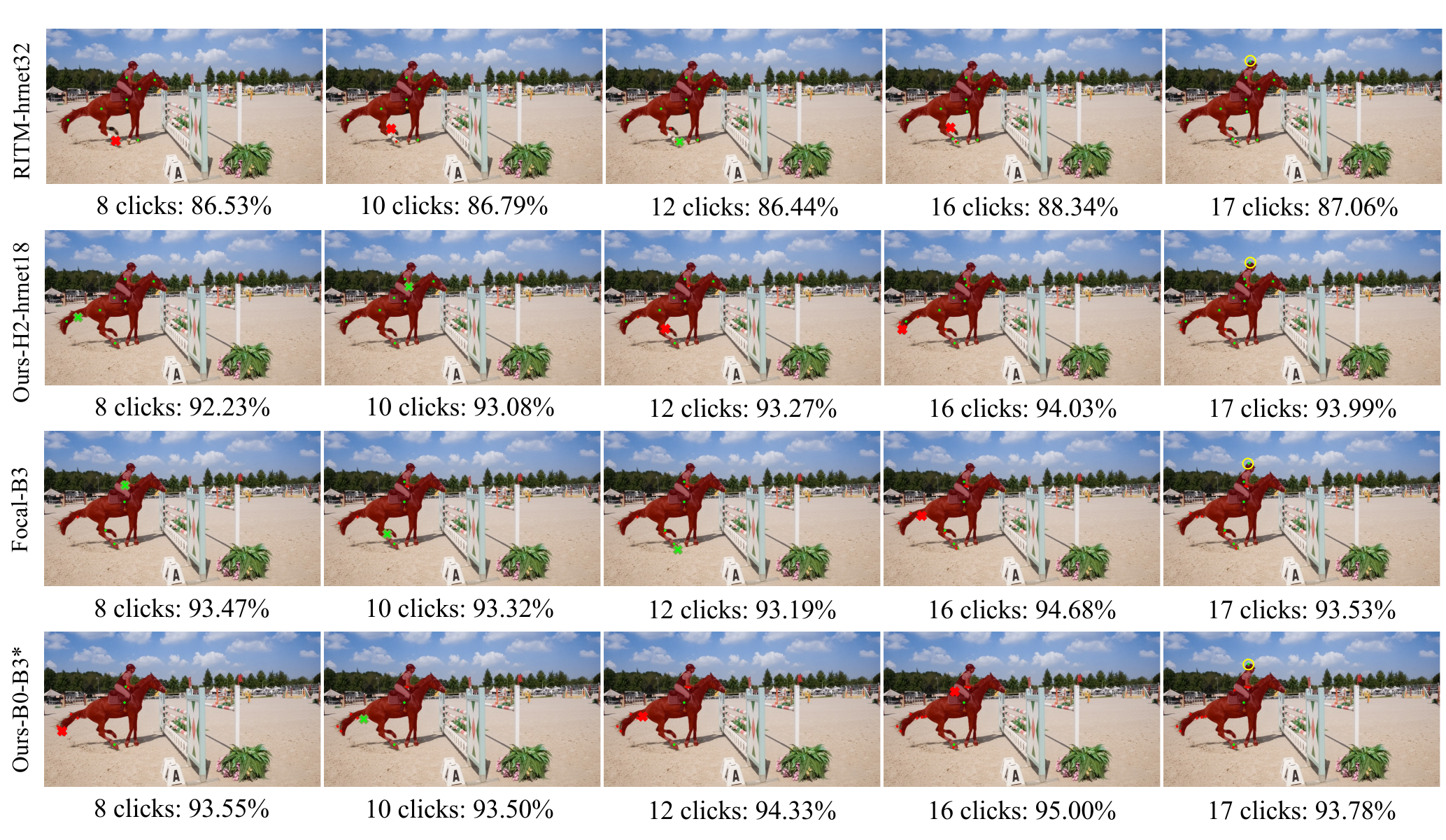}\\
\vspace{-10pt}\caption{Qualitative comparisons of robustness against misleading user guidance, where green/red dots denote positive/negative clicks. `B3' denotes segformerB3.  The false clicks and repetitive clicks are denoted with cross and circle, respectively.}\vspace{-6pt}
\label{fig:user intent visu}
\end{figure*}

\subsection{Qualitative Comparisons under Misleading User Guidance} 
To test the robustness of different models against misleading user guidance, in Fig.~\ref{fig:user intent visu}, we visualize the segmentation results of different models including our H2-hrnet18, our B0-segformerB3*, RITM-hrnet32~\cite{sofiiuk2021reviving} and Focal-segformerB3~\cite{focalclick}, under randomly generated misleading clicks. For each test sample, we first randomly select 5 clicks out of 20 clicks as the misleading clicks, while keeping the left 15 clicks as effective clicks. The 5 misleading clicks are either repetitive clicks or false clicks. All models are tested with the same sequence of misleading clicks for a fair comparison. The repetitive clicks and false clicks are denoted with circles and cross in Fig.~\ref{fig:user intent visu}, respectively. It can be observed that despite the perturbation from misleading clicks, our method is more robust than other methods. 


\end{document}